\documentclass[conference]{IEEEtran}
\IEEEoverridecommandlockouts
\usepackage{cite}
\usepackage{amsmath,amssymb,amsfonts}
\usepackage{algorithmic}
\usepackage{graphicx}
\usepackage{textcomp}
\usepackage{xcolor}
\usepackage{url} 
\usepackage{float}
\usepackage{booktabs} 
\usepackage{fancyhdr} 

\def\BibTeX{{\rm B\kern-.05em{\sc i\kern-.025em b}\kern-.08em
    T\kern-.1667em\lower.7ex\hbox{E}\kern-.125emX}}
\begin{document}


\pagestyle{empty}

\title{A Cloud-Based Hybrid Model for Real-Time Detection of BRTA-Approved Licence Plates Using YOLO Tiny and Haar Cascade\\


}

\author{
\IEEEauthorblockN{Debashis Kar Suvra}
\IEEEauthorblockA{
    \textit{Institute of Appropriate Technology} \\
    \textit{Bangladesh University of Engineering and Technology} \\
    Dhaka, Bangladesh \\
    0421292001@iat.buet.ac.bd
}
\and
\IEEEauthorblockN{Tahsina Farah Sanam}
\IEEEauthorblockA{
    \textit{Institute of Appropriate Technology} \\
    \textit{Bangladesh University of Engineering and Technology} \\
    Dhaka, Bangladesh \\
    tahsina@iat.buet.ac.bd
}
}

\maketitle

\begin{abstract}
Accurate vehicle license plate detection is essential for applications such as intelligent transportation systems, toll collection, parking management, and law enforcement. In Bangladesh, this task presents distinct challenges due to the complexity of localized license plates and environmental factors like lighting, occlusion, motion blur, and obstructions such as dirt or mud. These challenges often render conventional methods ineffective. This paper introduces a novel hybrid approach, combining the YOLO Tiny deep learning model with the Haar-Cascade classifier, for enhanced detection and localization of Bengali license plates. A key innovation of our system is the integration of a dynamic retraining pipeline, which allows the model to adapt to evolving real-world conditions. This retraining mechanism significantly boosts performance in low-confidence scenarios by continuously improving the model's accuracy as new data is encountered. Additionally, a publicly accessible dataset of BRTA-compliant license plates, captured under diverse and challenging conditions, has been developed to support this approach. Experimental results demonstrate that our approach not only achieves superior detection accuracy and computational efficiency over conventional models but also ensures consistent performance in resource-constrained environments, particularly in Bangladesh.
\end{abstract}


\section{Introduction}
Traditional license plate recognition (LPR) systems, designed primarily for recognizing English alphanumeric characters, face significant challenges when applied to region-specific standards \cite{cam2023vehicle}. The Bangladesh Road Transport Authority (BRTA) issues license plates incorporating Bengali script and unique designs in Bangladesh. These factors pose distinct challenges for conventional LPR systems, leading to frequent inaccuracies. Bengali script, known for its intricate characters and varying font styles, further complicates the recognition process, making it difficult for existing systems to perform effectively.

Little is known about the impact of these variations on existing recognition systems \cite{YOLOv5}. Most research relies on a single traditional approach \cite{Morphological}, which struggles to address real-world complexities. In addition to inconsistent lighting, occlusions from surrounding objects, and motion blur from fast-moving vehicles, the non-uniformity in vehicle types and license plate formats compounds the difficulty of recognition systems \cite{Challenging_env, Noisy}. In practical deployments, these environmental and design factors severely impact detection accuracy, underscoring the need for robust, localized solutions tailored to the distinctive characteristics of Bengali license plates and the unique variety of vehicle

To address these limitations, this paper introduces a hybrid approach that leverages the efficiency of YOLO Tiny v4, a deep learning-based object detection model, alongside the Haar-Cascade classifier for precise license plate localization. YOLO Tiny v4’s lightweight architecture\cite{wang2022trc}  facilitates real-time vehicle detection under diverse conditions, including long-range scenarios. Our approach optimizes computational efficiency by uniquely integrating Haar-Cascade for precise license plate localization within detected vehicle bounding boxes. Combined with a dynamic retraining pipeline, this hybrid method represents a key innovation, particularly in its adaptability to complex, low-confidence scenarios, ensuring consistent detection accuracy in resource-limited contexts. This layered approach significantly enhances both speed and accuracy compared to applying either method independently.

Most studies focus on either deep learning or traditional computer vision techniques \cite{pias2017bangladeshi, Localized}. Still, these approaches struggle to handle the intricacies of Bengali characters and the environmental challenges commonly encountered in real-world conditions and adapting to changes. Moreover, many existing models require significant computational resources and on-ground infrastructure installations \cite{amzad2021number}, making them impractical for real-time applications in resource-constrained environments like Bangladesh. In contrast, our system only requires an IP camera and a stable internet connection, making it a more feasible solution for real-time deployment in such environments.

A notable gap in current research is the lack of adaptive systems capable of continuously learning and improving \cite{transfer}. Our dynamic retraining pipeline addresses this gap by enabling the model to automatically adapt to new conditions in real-time, retraining itself on data from low-confidence detections. This continuous learning mechanism ensures the system maintains high accuracy, even in rapidly changing environments.

A major obstacle in Bengali license plate recognition is the lack of publicly available datasets, with most being limited in scope or inaccessible, significantly impeding research. Existing systems are often unable to handle real-world challenges like variable lighting, fast-moving vehicles, and occluded plates, reducing their practical effectiveness. 

Furthermore, this study contributes a new public dataset tailored specifically for Bengali license plates, featuring images captured under varied real-world conditions, including diverse lighting and motion speeds. This dataset aims to facilitate future research and pave the way for more effective LPR systems in regions like Bangladesh.

In conclusion, this paper not only addresses the technical challenges of recognizing BRTA-compliant Bengali license plates but also fills significant research gaps in the field of localized script recognition. By introducing a hybrid solution that combines YOLO Tiny v4 and Haar-Cascade, along with providing a publicly accessible dataset, this work lays the foundation for developing more efficient and accurate vehicle and license plate detection systems for complex, real-world environments.

\section{Literature review}


Real-time object detection has witnessed remarkable advancements in recent years, driven by integrating deep learning techniques. While traditional methods, such as Haar cascades and HOG(Histogram of Oriented Gradients), laid the foundation for early object detection systems, their limitations in handling complex scenes and multiple object classes paved the way for CNN-based approaches. Pioneering work introduced a robust framework for face detection using Haar-like features and AdaBoost, laying the foundation for subsequent research \cite{viola2001rapid}. However, the introduction of CNNs marked a pivotal turning point in object detection, enabling significant progress in accuracy and speed \cite{krizhevsky2012imagenet}. Region-based CNNs (R-CNN) proposed, and their variants played a crucial role in achieving real-time object detection \cite{girshick2014rich}. One-stage detectors, such as SSD \cite{liu2016ssd} and YOLO \cite{redmon2016you}, further accelerated the adoption of real-time object detection by achieving impressive speed and accuracy trade-offs. Recent advancements have been fueled by improvements in model architectures and the availability of large-scale annotated datasets. EfficientDet \cite{tan2020efficientdet} has demonstrated remarkable efficiency and accuracy, while domain-tailored models like YOLOv4-tiny \cite{bochkovskiy2020yolov4} underscore the adaptability of these systems. As the field continues to evolve, the balance between accuracy and computational efficiency remains a central challenge, driving ongoing innovations in this critical area of computer vision.


Computer vision (CV) is a field of artificial intelligence (AI) that enables computers and machines to derive meaningful information from digital images and videos. It involves developing algorithms to automatically extract and analyze visual features from photos and videos, such as objects, faces, and scenes. CV has a wide range of applications, including image search, facial recognition, video surveillance, medical imaging, and robotics \cite{guo2022attention}. On the other hand, cloud computing is a model for enabling ubiquitous, convenient, on-demand network access to a shared pool of configurable computing resources (e.g., networks, servers, storage, applications, and services) that can be rapidly provisioned and released with minimal management effort or service provider interaction \cite{cloud2011nist}. Cloud computing offers several advantages for CV applications, including scalability, cost-effectiveness, and accessibility.
CV applications can be very computationally demanding, especially when processing large volumes of images or videos. Cloud computing provides a scalable platform for CV applications, allowing users to access a large pool of computing resources on demand. This makes it possible to scale CV applications up or down as needed without investing in and managing their own hardware and software infrastructure. Cloud computing can help to reduce the cost of developing and deploying CV applications. This is because cloud providers offer various pay-as-you-go pricing options, which can help businesses avoid the upfront costs of purchasing and maintaining hardware and software.

\section{Proposed Methodology}

This approach combines deep learning and machine learning to create an efficient and reliable Bengali license plate recognition system. Using transfer learning on pre-trained models, a hybrid method of YOLOv4-tiny for vehicle detection and Haar-Cascade for plate localization, and real-time video analysis, we ensure robust performance across diverse conditions while maintaining computational efficiency for scalability.

\subsection{Data Collection \& Dataset Development} \label{AA}
This study's significant contribution is the creation of a custom dataset comprising both images and videos of vehicles displaying BRTA-compliant license plates, addressing the scarcity of Bengali license plate datasets. Assembling this data presented challenges due to the limited availability of such datasets. High-quality videos were captured using a Cruiser SE+ 4MP IP camera installed at the main gate of the Institute of Appropriate Technology (IAT) at the Bangladesh University of Engineering and Technology (BUET). The camera was positioned at an optimal height of 1.5 meters with a 6mm focal length, ensuring a clear view of passing vehicles. In addition to the fixed camera, supplementary images were collected using a mobile phone. Throughout the data collection process, ethical guidelines were strictly adhered to, ensuring both privacy and legal compliance, enhancing the dataset’s reliability and the trustworthiness of our system.


 \subsubsection{Diverse Lighting Conditions and Weather Variations}\label{subsubsec2}
Videos in the dataset were captured under varying lighting conditions, including bright sunlight, moderate daylight, and low-light night scenes. This diversity ensures that the model maintains robustness and accuracy across different lighting environments, whether during the day or at night. Additionally, images captured in various weather conditions, such as clear skies and rain, further provide comprehensive testing for the model's performance in real-world scenarios.

 \subsubsection{Diverse Vehicle Types}\label{subsubsec3}
The dataset includes images of different vehicle types commonly seen on Bangladeshi roads, such as cars, motorcycles, auto-rickshaws, vans, and buses. This variation ensures that the model generalizes well across different vehicle classes, broadening its applicability on diverse road conditions.

\subsection{YOLOv4 tiny for Vehicle Detection}
For vehicle detection, the system employs the YOLOv4-tiny model, pre-trained on the MS COCO dataset, which contains a wide range of object classes, including vehicles. The model efficiently detects objects in real time by predicting bounding boxes and confidence scores for each grid cell. A confidence threshold filters out low-confidence detections, ensuring reliable results during object detection

 For each grid cell, YOLOv4-tiny predicts bounding boxes with their coordinates (\(x, y, w, h\)) and confidence scores (\(C\)):

\[ B = (x, y, w, h, C) \]

Where:
\begin{itemize}
  \item \( (x, y) \) are the center coordinates of the bounding box relative to the grid cell,
  \item \( (w, h) \) are the width and height of the bounding box,
  \item \( C \) is the confidence score indicating the probability of the box containing an object.
\end{itemize}

\[ P(\text{class} | B) = \frac{\exp(B \cdot W_c)}{\sum_{i=1}^N \exp(B \cdot W_i)} \]

Where:
\begin{itemize}
  \item \( P(\text{class} | B) \) is the probability of the bounding box containing a particular class,
  \item \( B \) is the concatenation of bounding box coordinates and confidence score,
  \item \( W_c \) is the weight vector for the class \( c \),
  \item \( N \) is the total number of classes.
\end{itemize}

A confidence score threshold is applied during post-processing, and bounding boxes with confidence scores below this threshold are discarded. This helps filter out specific detections.

\subsection{Haar-Cascade for License Plate Localization}

The Haar Cascade Classifier \cite{opencvdoc} is a machine-learning object detection method used for license plate detection. It is a pre-trained model \cite{ashrafee2022real} generated from a dataset containing positive and negative samples. Positive samples consist of annotated license plate regions of BRTA-aligned license plates from all over Bangladesh, while negative samples represent areas without license plates. The classifier undergoes training to learn patterns that distinguish between the positive and negative samples.

Mathematically, the training process involves optimizing a set of weights and thresholds in the classifier. Let's denote the positive samples as \(P\) and the negative samples as \(N\). For each feature in the Haar-like feature set, the classifier learns a weight (\(w\)) and a threshold (\(t\)). The final decision function can be represented as:

\[ H(x, y) = \sum_{i} w_i \cdot f_i(x, y) - t \]

Where:
\begin{itemize}
  \item \(H(x, y)\) is the decision function,
  \item \(w_i\) is the weight for feature \(i\),
  \item \(f_i(x, y)\) is the value of feature \(i\) at position \((x, y)\),
  \item \(t\) is the threshold.
\end{itemize}

During training, the weights and thresholds are adjusted to minimize classification errors, effectively discriminating between positive and negative samples \cite{ashrafee2022real}.

For Haar-like features, each feature is a simple rectangular filter, and the value of the feature is the difference between the sum of pixel intensities in the white and black rectangles. The selection and arrangement of these features contribute to the classifier's ability to recognize specific patterns, such as those found in license plates.

After successfully detecting the license plate, our technology meticulously checks the data to ensure the utmost confidence and accuracy before saving it for storage. The system examines the confidence rates associated with the recognized license plates when a vehicle is spotted. It chooses the highest confidence rate which is more than 80\%, ensuring the recorded data is as accurate and dependable as possible.

\subsection{Fine Tuning, Transfer learning \& Hybrid Approach}

The pretraining process begins with a YOLOv4-Tiny model that has been trained on a large dataset. In this model, the earlier layers—responsible for feature extraction (e.g., detecting edges, textures, and shapes) are retained during the transfer learning process \cite{carrasco2021t}. The final layers, specifically the detection head layers, which include the output layer responsible for predicting bounding boxes \cite{feng2022benchmark}, object classes, and confidence scores, are fine-tuned using a custom dataset of Bengali license plates.

During transfer learning, only the detection head layers are retrained using our dataset, allowing the model to specialize in identifying Bengali license plates under real-world conditions. The early layers remain frozen to preserve their learned features, ensuring computational efficiency. Anchors were carefully selected based on the aspect ratio of license plates, and we used a batch size of 16, which was suitable given the memory limitations of our hardware. The learning rate was set to 0.001 to balance convergence speed and stability. Training began with pre-trained weights, leveraging the knowledge from existing datasets while adapting the model to the unique characteristics of Bengali script and varying environmental conditions.

\begin{figure}[htbp]
    \centering
    \includegraphics[width=\linewidth]{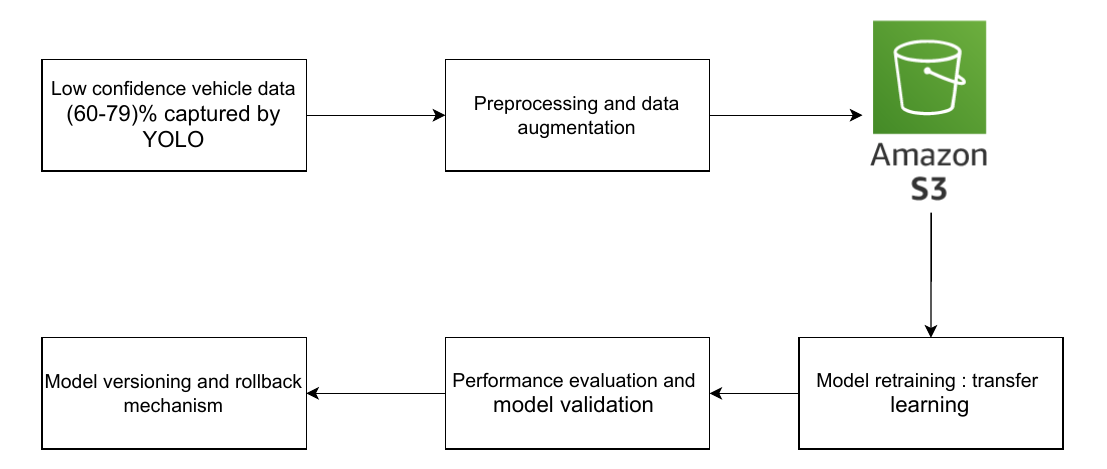} 
    \caption{Model retraining pipeline}
    \label{fig:retraining}
\end{figure}

\begin{figure}[htbp]
    \centering
    \includegraphics[width=\linewidth]{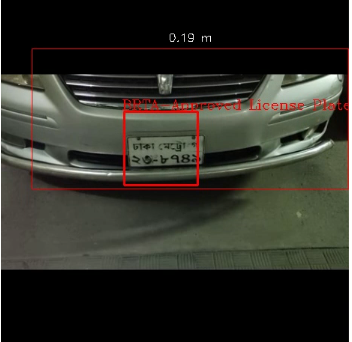} 
    \caption{Detection of BRTA-approved license plate using hybrid model}
    \label{fig:Detection}
\end{figure}

\subsubsection{Efficiency Strategies}

This dynamic approach adapts to varying scenarios, focusing computational resources on regions where vehicles are closer to the camera around 1 meter or more. It ensures that only relevant regions undergo subsequent processing, mitigating unnecessary computations on distant or irrelevant objects. This adaptability increases accuracy, particularly in scenarios with varying distances between the camera and detected vehicles. After YOLOv4 efficiently identifies and localizes vehicles in the entire frame, the subsequent processing for license plate recognition is concentrated solely on the extracted ROI. This targeted application of computational resources is defined by:

\[ R_{\text{Processed}} = \{B_i \mid B_i \in R_{\text{ROI}}\} \]

Computational resources are conserved by employing YOLOv4 for initial vehicle detection and restricting subsequent processing to the ROI. This approach optimizes processing speed and minimizes the likelihood of false positives in license plate recognition. The result is a more resource-efficient system with enhanced responsiveness.

\subsubsection{Cascade Approach and Elimination of Irrelevant Regions}

The cascade approach involving YOLOv4 followed by the Haar-cascade-based License Plate detector ensures a sequential and targeted examination of the video frame. This process avoids unnecessary computations on irrelevant regions, contributing to increased accuracy in Bangla license plate detection.

A hierarchical processing approach is applied, where YOLOv4 performs broad vehicle detection, followed by the Haar Cascade focusing on license plates. This sequential approach ensures accurate detection without unnecessary computation, boosting system performance.

\subsection{System Integration and Performance Evaluation}

To establish a scalable and efficient testing environment, we utilized cloud technology. A Flask server was deployed on a cloud platform to host our hybrid model, providing a flexible and accessible interface for real-time testing. Real-time video streams were ingested using the Real-Time Streaming Protocol (RTSP), simulating live traffic scenarios and allowing for comprehensive evaluation of the model's performance.

Upon receiving a video frame, the YOLOv4 Tiny model was employed to detect vehicles within the scene. Once a vehicle was successfully identified, the Haar-Cascade classifier was activated as a "wake-up" mechanism to focus on the region of interest. The Haar-Cascade classifier was then used to localize the BRTA-aligned license plate within the detected vehicle region. The final output of the system consisted of the detected vehicle and its corresponding license plate.

This testing setup provided a robust and scalable environment for evaluating the performance of our hybrid model in a real-world context.

\begin{table*}[htbp]
\centering
\caption{Performance Metrics of Hybrid Model with Data Augmentation Techniques}
\label{tab:Data_Augmentation_Performance_Lower}  
\begin{tabular}{lccccc}
\toprule
\textbf{Data Augmentation Technique} & \textbf{Precision (\%)} & \textbf{Recall (\%)} & \textbf{F1 Score (\%)} & \textbf{Accuracy (\%)} & \textbf{FPS} \\
\midrule
No Data Augmentation & 76 & 74 & 75 & 72 & 10 \\
Scaling & 80 & 78 & 79 & 76 & 19 \\
Contrast Adjustment & 75 & 72 & 73 & 69 & 18 \\
Dynamic Range Enhancement & 82 & 80 & 81 & 78 & 16 \\
Noise Reduction & 79 & 77 & 78 & 75 & 15 \\
\bottomrule
\end{tabular}
\end{table*}

\section{Experimental Study}

In this section, we present the experimental evaluation of the proposed hybrid approach, combining YOLO Tiny v4 and the Haar-Cascade classifier for license plate detection and recognition. The performance metrics evaluated include precision, recall, F1 score, accuracy, and frames per second (FPS). We also assessed the impact of various data augmentation techniques. The experiments used a dataset of BRTA-compliant Bengali license plates, captured under real-world conditions with challenges like varying lighting, occlusions, and vehicle motion.

\subsection{Experimental Setup}

In developing our method, we prioritized real-time inference speed while ensuring accessibility for a wider audience. While some approaches, such as \cite{laroca2018robust} emphasize high-speed processing using powerful GPUs like the NVIDIA Titan XP, these resources are often out of reach for many.
To replicate real-world deployment conditions and ensure accurate performance metrics, we conducted all experiments within the AMD Ryzen Setup (ARS). ARS features an AMD Ryzen 3 2200G processor with Radeon Vega Graphics, operating at 3.50 GHz, equipped with 16GB of RAM. Ubuntu 22.04 LTS, a stable and widely used Linux distribution, was selected as the operating system to ensure compatibility.  We collected data using a Dahua IMOU CRUISER SE+4MP IP camera with a 20 Mbps internet connection. Test videos were preprocessed to a standardized resolution of 464x464 pixels for consistent inference times and performance comparisons. Initial experiments were conducted in the AMD Ryzen Setup (ARS) to ensure real-time inference speed, followed by a cloud-based deployment using a Flask server for scalable real-world testing. Real-time video streams were processed through RTSP for both environments to ensure consistency.

The hybrid model, using pre-trained weights and configurations, was tested with live footage from a Dahua IMOU CRUISER SE+4MP IP camera streaming via RTSP. A confidence threshold of 0.8 ensured reliable vehicle detection, while a 0.1 non-maximum suppression threshold reduced redundant bounding boxes for accurate results~\cite{ashrafee2022real, open_model_zoo}.

For license plate detection, the application employs a Haar Cascade classifier with a scaling factor of 1.05, a minimum of 5 neighbors, and a minimum plate size of (60, 80) pixels. The processed frames are streamed through a Flask web server.

\subsubsection{Data Augmentation Techniques}

Data augmentation plays a crucial role in improving the robustness and generalization of machine learning models \cite{maharana2022review}. Table~\ref{tab:Data_Augmentation_Performance_Lower} demonstrates the effects of various augmentation techniques on the performance metrics of a hybrid model. Techniques like scaling adjust image dimensions, contrast adjustment modifies lighting variations, and dynamic range enhancement improves visibility under different lighting conditions. Noise reduction further refines the signal by minimizing unwanted variations. These strategies significantly impact precision, recall, F1 score, accuracy, and FPS, underscoring their importance in enhancing the model's performance for tasks like license plate recognition.

\subsection{Result and Comparative Discussion}

In the experimental evaluation, we used the YOLOv4-Tiny model, pre-trained on the MS COCO dataset for vehicle detection. Due to the limited size of our dataset, only the final layers for bounding box prediction and classification were fine-tuned, while the pre-trained layers for feature extraction were kept frozen to retain general detection capabilities.

\begin{figure}[htbp]
    \centering
    \includegraphics[width=\linewidth]{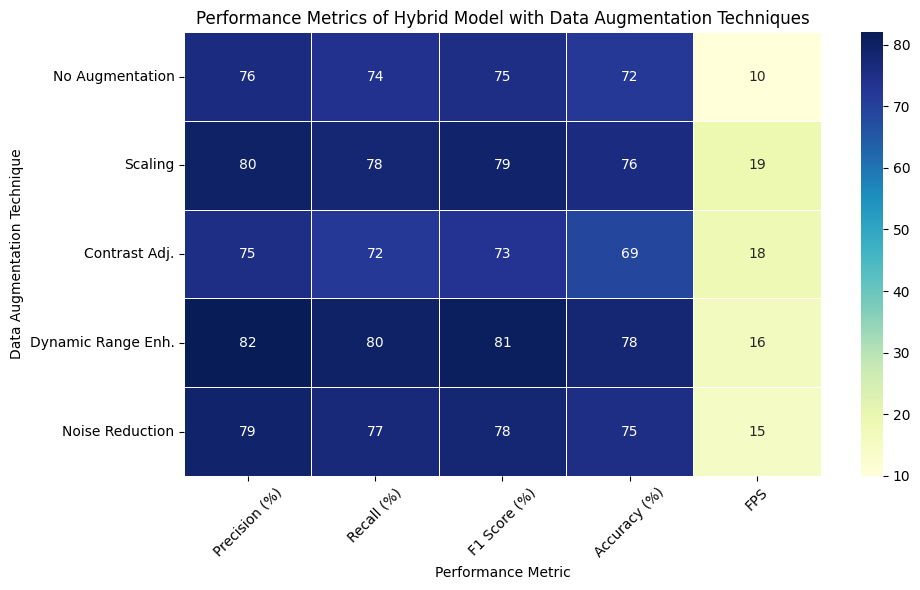} 
    \caption{The performance analysis for models with and without data augmentation techniques}
    \label{fig:performance}
\end{figure}

As shown in Table \ref{tab:Data_Augmentation_Performance_Lower}, the performance metrics of the hybrid model with various data augmentation techniques demonstrate significant improvements in precision, recall, F1 score, accuracy, and frames per second (FPS). The fine-tuning focused on improving performance under specific conditions like lighting variations and occlusions. Although our dataset was insufficient for full retraining, this selective adjustment enhanced the model's adaptation to localized challenges. The hybrid model was tested with live RTSP streams, ensuring real-time performance while balancing computational efficiency and accuracy. The performance metrics of the license plate recognition models are summarized in Table \ref{tab:performace_metrics}, highlighting the precision, recall, F1 score, accuracy, and frames per second (FPS) for each model. Figure \ref{fig:performance} illustrates the performance analysis for models with and without data augmentation techniques.

\begin{table*}[htbp]
\centering
\caption{Performance Metrics of License Plate Recognition Models}
\label{tab:performace_metrics}  
\begin{tabular}{lccccc}
\toprule
\textbf{Model} & \textbf{Precision (\%)} & \textbf{Recall (\%)} & \textbf{F1 Score (\%)} & \textbf{Accuracy (\%)} & \textbf{FPS} \\
\midrule
YOLO Tiny v4 & 93 & 91 & 92 & 89 & 40 \\
Haar-Cascade & 75 & 70 & 73 & 70 & 19 \\
YOLO Tiny v4 + Cascade (Hybrid) & 94 & 93 & 93 & 91 & 30 \\
\bottomrule
\end{tabular}
\end{table*}

\section{Conclusion}

This study introduces a novel hybrid approach to tackle the challenges of detecting and recognizing BRTA-compliant Bengali license plates, which are unique due to their script and diverse environmental conditions. By integrating the speed and efficiency of YOLO Tiny v4 for vehicle detection with the precision of the Haar-Cascade classifier for license plate localization, the proposed system significantly improves performance over traditional methods. The model achieved high accuracy, with a precision of 94\%, recall of 93\%, and an F1 score of 93\%, demonstrating its effectiveness in real-world applications. Our solution addresses the specific challenges of recognizing Bengali scripts in varied environmental conditions, such as lighting and motion blur, making it well-suited for resource-constrained real-time applications. A key contribution of our system is the integration of a dynamic retraining pipeline, which allows the model to continuously adapt to changing conditions, significantly improving performance in low-confidence detection scenarios.

In addition, the use of data augmentation techniques, particularly dynamic range enhancement, further strengthened the model’s resilience to challenging conditions such as lighting variations and occlusions. This improvement ensures the system’s reliability in real-world deployments, where traditional methods often struggle. The creation of a dedicated public dataset for Bengali license plates also contributes to advancing the field, providing a valuable resource for further research in non-Latin script recognition.

In summary, the hybrid framework enhances both the accuracy and computational efficiency of license plate detection and recognition, offering a robust solution tailored to the specific needs of regions like Bangladesh. This work lays the groundwork for future research in developing even more advanced systems by integrating additional deep learning techniques and refining data augmentation strategies, ensuring adaptability and performance in diverse and complex environments.

\bibliographystyle{IEEEtran}
\bibliography{strings} 

\mbox{} 

\end{document}